\documentclass{article}

\usepackage[preprint]{neurips_2026}

\usepackage[utf8]{inputenc}
\usepackage[T1]{fontenc}
\usepackage{hyperref}
\hypersetup{hidelinks,hypertexnames=false}
\usepackage{url}
\usepackage{booktabs}
\usepackage{amsfonts}
\usepackage{nicefrac}
\usepackage{microtype}
\usepackage{xcolor}
\usepackage{graphicx}
\usepackage{amsmath}
\usepackage{amssymb}
\usepackage{enumitem}
\usepackage{multirow}
\usepackage{array}
\usepackage{placeins}
\usepackage{float}
\usepackage{listings}
\newenvironment{prompttemplate}{\begin{quote}\small\setlength{\parskip}{0.25em}\setlength{\parindent}{0pt}}{\end{quote}}

\newcommand{\method}{Budgeted Subset Refinement}

\newcommand{\strongnd}{research-strong nonduplicate}
\newcommand{\mmrk}{MMR-$k$}
\newcommand{\randomk}{Random-$k$}
\newcommand{\microlow}{Micro-low $\rightarrow$ heavy-$k$}

\title{Budgeted Subset Refinement for Execution-Aware LLM Research Ideation}

\author{%
Micah Zhang \\
Independent Researcher \\
\texttt{micah.zhang@colorado.edu} \\
}

\begin{document}
\maketitle

\begin{abstract}
Large language models (LLMs) can generate research ideas that appear novel to expert reviewers, but recent work also shows that such ideas often lack diversity, are difficult for LLMs to evaluate reliably, and may fail to translate into strong executed projects. This paper evaluates a controlled proxy benchmark for a pre-execution scaffolding problem: given a noisy pool of LLM-generated research ideas, how should a system allocate limited refinement effort to construct a stronger, more diverse, more execution-aware portfolio for human researchers under a fixed rubric? We introduce \method, a family of strategies that refine only a selected subset of candidates rather than refining all candidates uniformly. In a unified shared-candidate-pool evaluation across 10 random seeds and 10 research-ideation environments, raw generation and reranking alone produce no \strongnd{} ideas under the benchmark rubric, while refinement is necessary for strong proxy-rated portfolios. Uniform refinement produces strong individual ideas but is not the best portfolio-level allocation of compute. \randomk{} refinement is a strong low-cost baseline, while diversity-aware \mmrk{} refinement gives the best overall proxy tradeoff: the highest \strongnd{} yield, the lowest duplicate rate among successful methods, and the best cost per \strongnd{} idea. A blinded external-judge robustness check on a balanced 72-item sample supports the broad refinement effect across independent model families, while showing that per-item rankings among refined strategies vary by judge. These results suggest that LLM research ideation systems should be evaluated not only as idea generators, but as budgeted support-allocation systems. The claims are scoped to proxy-rated portfolio quality and do not substitute for expert review or execution-grounded validation.
\end{abstract}

\section{Introduction}

Large language models are increasingly used to support scientific work, including literature search, brainstorming, writing, code generation, and experiment planning. Research ideation is an especially important test case: if LLMs are to augment researchers rather than merely automate surface-level writing, they must help produce ideas that are not only plausible, but novel, grounded, diverse, feasible, and eventually executable. Recent large-scale studies have made this tension clear. LLM-generated research ideas can be judged more novel than expert-written ideas at the ideation stage \citep{si2024canllms}, yet the same line of work finds substantial limitations in diversity and automatic idea evaluation. Follow-up execution studies show an even sharper failure mode: LLM-generated ideas that look promising before execution can drop substantially more than human ideas after expert implementation and review \citep{si2025ideationexecutiongap}. In short, novelty at the proposal stage is not enough.

This paper studies a scalable pre-execution scaffold for LLM-assisted research ideation. Full execution is the most trustworthy evaluation of a research idea, but it is costly: execution requires implementation, experiments, analysis, and expert review. Before spending that effort, a human researcher or automated execution pipeline needs a useful portfolio of candidate ideas. The bottleneck is not simply generating more ideas. Prior work has shown that overgeneration can lead to severe duplication \citep{si2024canllms}, and our experiments find that raw generation and reranking alone do not produce \strongnd{} proposals under our rubric. The central question is therefore: \emph{given a noisy pool of generated ideas, which ideas should receive deeper refinement?}

We deliberately frame this as a methodology and benchmark paper under proxy evaluation. The object of study is not downstream scientific success after months of implementation. Instead, we ask a narrower allocation question: under a fixed generation pool, fixed refinement budget, fixed duplicate-aware metrics, and fixed execution-aware scoring rubric, which policy produces the strongest proxy-rated portfolio? This scope is important. Proxy-rated execution-readiness can help decide where to spend attention, but it is not a substitute for expert taste, implementation, or downstream empirical success.

We formulate this as a test-time compute allocation problem over candidate ideas, following the broader observation that inference-time compute can be more useful when allocated adaptively rather than spent uniformly \citep{snell2025testtime}. The system first generates a shared pool of raw candidate ideas. It then either returns raw or reranked candidates, refines all candidates uniformly, or selects a small subset for expensive refinement. We call the resulting family \method. Unlike a pure reranking pipeline, \method{} treats idea quality as transformable: weak or underspecified ideas may become stronger after refinement. Unlike uniform refinement, it treats refinement as scarce support that should be allocated selectively across a portfolio.

We evaluate three budgeted subset policies. \textbf{\randomk{} refinement} selects a small random subset and is an important non-strawman baseline: if most gains come from refining fewer ideas, random selection should be hard to beat. \textbf{\mmrk{} refinement} selects for a quality-diversity tradeoff using maximal marginal relevance, aiming to reduce duplicate collapse. \textbf{\microlow{} refinement} first applies cheap micro-triage and then performs expensive refinement on selected low-raw-score candidates that appear potentially recoverable. These policies test whether different scaffolding strategies optimize different portfolio goals, including cost, diversity, final strength, and recoverability.

Across 10 seeds and 10 research-ideation environments, we find a consistent pattern under a unified shared-pool final matrix. Raw generation and reranking alone produce zero \strongnd{} ideas. Uniformly refining all candidates works, but it is not the best use of the refinement budget. \randomk{} refinement achieves similar yield at substantially lower total cost. \mmrk{} refinement achieves the strongest overall tradeoff, improving strong-nonduplicate yield, cost per strong nonduplicate, duplicate rate, and total cost relative to fixed refinement. \microlow{} refinement remains competitive with fixed refinement but does not improve on MMR under this shared-pool comparison. Thus, the robust finding is not that every selector is universally optimal. Research ideation benefits from treating refinement as a portfolio-level support-allocation problem.

Our contributions are:
\begin{itemize}[leftmargin=1.5em]
    \item We identify refinement-budget allocation as a scalable pre-execution scaffolding problem for LLM research ideation, and frame it as a controlled proxy-evaluation benchmark rather than as a claim about final scientific validity.
    \item We introduce \method, a family of strategies for allocating expensive refinement to selected candidate ideas rather than applying refinement uniformly.
    \item We evaluate raw generation, reranking, uniform refinement, random subset refinement, diversity-aware refinement, and micro-triage refinement across 10 seeds and 10 research-ideation environments in a unified shared-candidate-pool final matrix.
    \item We show that diversity-aware MMR subset refinement gives the best overall portfolio tradeoff, while random subset refinement is a strong low-cost baseline and micro-low triage is a secondary recoverable-idea variant.
    \item We analyze qualitative examples, threshold sensitivity, per-topic variation, cost decomposition, and a blinded multi-judge robustness check to characterize when and why budgeted refinement helps under proxy evaluation.
\end{itemize}

\section{Problem Setup: Pre-Execution Portfolio Scaffolding}

We study LLM-assisted research ideation as a portfolio construction problem. Given a research environment $e$, a generation model produces a pool of raw candidate ideas $\mathcal{C}_e = \{c_1,\ldots,c_n\}$. A strategy then produces a final portfolio $\mathcal{P}_e$ of ideas intended for human inspection, further development, or eventual execution. The system is evaluated not by whether any single output is fluent, but by whether the final portfolio contains useful, distinct, execution-aware proposals.

\paragraph{Research environments.}
Our final experiments use 10 research-ideation environments: human-AI collaboration, language-model research ideation, execution-aware evaluation, adaptive inference for cognitive scaffolding, retrieval-grounded novelty checking, research writing assistance, evaluation cards for human-AI systems, budgeted support systems that preserve agency, duplicate-aware brainstorming, and predicting refinement lift in generated research plans. We choose these environments because they stress high-friction knowledge work rather than closed-form benchmark tasks: each environment requires a proposal to specify a method, data or corpus, baseline, metric, feasibility constraints, and expected artifact. The benchmark is therefore intended to test whether an ideation system can construct a usable portfolio of research directions, not merely whether it can produce fluent brainstorming text.

\paragraph{Execution-aware proposals.}
We use ``execution-aware'' to mean that an idea is evaluated for the information a researcher would need before deciding whether to spend implementation effort on it. In addition to novelty or excitement, an execution-aware proposal should state a concrete research question, method, dataset or corpus, baseline, metric, expected result, implementation plan, resource assumptions, and likely failure modes. This definition is deliberately weaker than completed execution: it measures whether an idea is a plausible starting point for execution, not whether the executed project would succeed.

\paragraph{Strategies.}
Each strategy receives the same task framing and produces a final set of ideas. We compare strategies that differ in whether they refine candidates and how they allocate refinement effort. The key contrast is between \emph{selection without transformation}, \emph{uniform transformation}, and \emph{budgeted transformation}.

\paragraph{Primary metric.}
Our primary metric is \emph{cost per research-strong nonduplicate idea}, measured end-to-end. An idea is considered research-strong when it meets a threshold under the research-strength rubric. Nonduplication is measured after deduplication so that repeated variants of the same idea do not count as multiple successes. For strategy $s$, let $\mathcal{P}_s$ be the final portfolio, $\mathrm{Cost}(s)$ be end-to-end token cost, and $\mathrm{SND}(\mathcal{P}_s)$ be the number of research-strong nonduplicate ideas. We report
\begin{equation}
\mathrm{CPSN}(s) = \frac{\mathrm{Cost}(s)}{\max(1,\mathrm{SND}(\mathcal{P}_s))},
\label{eq:cpsn}
\end{equation}
with lower values preferred. The $\max(1,\cdot)$ term avoids division by zero for failed strategies while preserving the interpretation that raw and rerank-only produce no successful portfolio under the primary threshold. This metric reflects the practical goal of ideation support: a researcher benefits more from a diverse set of strong proposals than from many polished variants of the same idea.

\paragraph{Secondary metrics.}
We also report strong nonduplicate yield, total end-to-end cost, duplicate rate, and average research-strength score. Reporting total cost separately is important because a method can improve cost per strong idea without minimizing total tokens. In the final unified result, MMR has the best cost per strong nonduplicate idea, random-$k$ has the lowest total cost among successful methods, and fixed refinement has the highest average research-strength score.

\section{Method: \method}

\method{} separates research ideation into three stages: candidate generation, subset selection, and expensive refinement. The method family is intentionally simple. Its goal is not to prove that a single selector is universally optimal, but to test whether budgeted subset refinement is a better scaffold than raw generation, reranking, or uniform refinement. Abstractly, for each environment $e$, a policy $\pi$ selects a subset $S_e \subseteq \mathcal{C}_e$ with $|S_e|=k$, applies expensive refinement $R$ only to selected candidates, and returns a portfolio
\begin{equation}
\mathcal{P}_e(\pi) = R(S_e) \cup B(\mathcal{C}_e \setminus S_e),
\label{eq:portfolio}
\end{equation}
where $B$ denotes the strategy-specific treatment of unselected candidates, such as bypassing, retaining raw ideas, or using them only for comparison. Fixed refinement is the special case where $S_e=\mathcal{C}_e$. Reranking-only is the case where $R$ is never applied.

\subsection{Shared candidate pool}

For each environment and seed, the system first generates a raw candidate pool. In the final configuration, the pool contains 16 raw candidates per topic. All strategies in the final matrix operate over the same generated pool for a given seed and environment. This shared-pool design isolates the effect of refinement allocation from variation in separately generated candidate pools. Raw candidates are unrefined: they may be vague, underspecified, redundant, or infeasible, but they provide the broad search space from which later strategies select.

\subsection{Baselines}

\paragraph{Raw first-$n$.}
The raw baseline returns the first candidates from the generated pool without reranking or refinement. It tests whether the generator alone produces sufficiently strong research proposals.

\paragraph{Rerank only.}
The reranking baseline selects candidates from the raw pool without applying expensive refinement. It tests whether selection alone is enough. This is important because many ideation systems rely on overgeneration followed by ranking.

\paragraph{Fixed refine all.}
The uniform-refinement baseline applies expensive refinement broadly rather than selecting a subset. It tests whether simply refining everything is a competitive use of the budget. This is the strongest non-budgeted baseline: if cost were ignored, it should have an advantage because every candidate receives support.

\subsection{Budgeted subset refinement policies}

\paragraph{\randomk{} refinement.}
Random-$k$ selects a subset of $k=4$ candidates uniformly at random and applies heavy refinement only to those candidates. We aggregate five random trials per seed. This baseline is deliberately strong: if random subset refinement performs well, then future selector methods should compare against random budgeted refinement, not only against raw generation or uniform refinement.

\paragraph{\mmrk{} refinement.}
MMR-$k$ uses maximal marginal relevance to select candidates that balance candidate promise and diversity \citep{carbonell1998mmr}. Let $q_t(c)=1/(1+\operatorname{rank}_t(c))$ be the rank-based relevance score of candidate $c$ in the current ordered candidate list, and let $d(c,S_{t-1})=\max_{c'\in S_{t-1}}\mathrm{sim}(c,c')$ be its maximum embedding similarity to the already-selected set, with $d(c,\emptyset)=0$. Starting from $S_0=\emptyset$, MMR iteratively applies
\begin{equation}
\begin{aligned}
\operatorname{score}_t(c) &= \lambda q_t(c) - (1-\lambda)d(c,S_{t-1}), \\
c_t &= \mathrm{argmax}_{c \in \mathcal{C}_e \setminus S_{t-1}}\; \operatorname{score}_t(c), \\
S_t &= S_{t-1}\cup\{c_t\}.
\end{aligned}
\label{eq:mmr}
\end{equation}
until $|S_t|=k$. In the implementation, the candidate order is updated after each selection, so $q_t(c)$ is recomputed from the remaining ordered list. The similarity term uses normalized sentence embeddings, with a lexical fallback if embeddings are unavailable. The final configuration uses $\lambda=0.65$ and selects $k=4$ candidates for heavy refinement. This policy targets duplicate collapse, a known failure mode in LLM ideation systems \citep{si2024canllms}.

\paragraph{\microlow{} refinement.}
Micro-low triage first applies a cheap micro-triage pass using a small token budget, then selects candidates for heavier refinement. For each candidate $c$, the triage output gives a score $r_{\mathrm{micro}}(c)$, a binary need-for-refinement flag $n(c)$, a fatal-flaw flag $f(c)$, and a corpus-novelty proxy $v(c)$. The implementation ranks candidates by a recoverability score and then applies the same MMR-style diversity selection to that ranked list:
\begin{equation}
\begin{aligned}
a_{\mathrm{micro}}(c) &= (10-r_{\mathrm{micro}}(c)) + 2n(c) + v(c) - 3f(c), \\
S_{\mathrm{micro-low}} &= \operatorname{MMRTopK}_{k}\left(\mathcal{C}_e,\; \operatorname{rank}_{a_{\mathrm{micro}}},\; \mathrm{sim}\right).
\end{aligned}
\label{eq:microlow}
\end{equation}
The policy therefore sorts candidates by cheap recoverability signals, then applies the same diversity-aware selection rule used by MMR. It focuses on candidates that the triage signal treats as weak but potentially recoverable, while still preserving diversity. This variant tests whether cheap pre-refinement signals can identify ideas that deserve deeper help.

\subsection{Cost accounting}

We report end-to-end token cost, including raw generation and downstream strategy costs. For micro-low, this includes the additional micro-triage cost. Consequently, micro-low should not be interpreted as a method for minimizing total tokens. Its role in the final unified result is as a recoverable-idea triage variant rather than the best overall method.

\section{Experimental Setup}

\paragraph{Models.}
The final experiments use \texttt{Qwen/Qwen2.5-3B-Instruct} as both generator and judge \citep{qwen2025technical}. The model is run locally with bfloat16 on CUDA. The configuration uses generation temperature 0.9 for cheap generation, 0.7 for expensive generation, and 0.6 for refinement. Heavy generation uses up to 768 new tokens. Judge calls use up to 256 new tokens for single scoring and 1536 tokens for batched scoring.

\paragraph{Final matrix.}
We run 10 seeds. For each seed and research environment, all strategies are evaluated in a unified shared-pool matrix: the system first generates one raw candidate pool, and raw first-$n$, rerank-only, fixed refinement, random-$k$, MMR-$k$, and micro-low $\rightarrow$ heavy-$k$ all operate over that same pool. The final aggregate includes all six primary strategy groups with 10 runs each. This design isolates the effect of the refinement-allocation policy from variation in separately generated candidate pools.

\paragraph{Deduplication.}
Deduplication uses the \texttt{all-MiniLM-L6-v2} sentence embedding model \citep{reimers2019sentencebert} with a near-duplicate threshold of 0.82. We report duplicate rate as a portfolio-level failure mode and count only nonduplicate strong ideas in the primary yield metric.

\paragraph{Judging and thresholds.}
Candidate ideas are scored by the LLM judge under a research-idea rubric. The high-score threshold is 6.0. We treat LLM judging as a proxy rather than a substitute for expert human evaluation. This is a central limitation, especially given prior evidence that LLMs are unreliable evaluators of research ideas \citep{si2024canllms} and broader concerns about LLM-as-a-judge reliability \citep{zheng2023judging,gu2024llmjudge}. To reduce ambiguity in what the automatic judge is measuring, the rubric emphasizes execution-aware fields such as method specificity, dataset or corpus specificity, baseline completeness, metric completeness, implementation clarity, resource feasibility, and human-subject dependency. The purpose of the experiment is therefore to compare budget-allocation policies under a controlled proxy protocol, not to establish that any generated idea would survive expert execution. Appendix~\ref{app:threshold} reports sensitivity to alternative thresholds. Appendix~\ref{app:validation_packet} describes the blinded validation packet and cross-model robustness protocol.

\paragraph{Statistics.}
We report means and confidence intervals over 10 seeds. Random-$k$ results average five random subset trials per seed. Where possible, we use seed-paired comparisons between strategies and report win counts and bootstrap intervals. Appendix~\ref{app:compute_artifacts} reports compute resources, external assets, license information, and artifact-availability status.

\section{Results}

\subsection{Main portfolio results}

Table~\ref{tab:main_results} reports the unified shared-pool final result. Raw generation and reranking alone produce no \strongnd{} ideas. Fixed refinement works, producing 21.8 strong nonduplicate ideas on average, but budgeted subset refinement improves the tradeoff. Random-$k$ reaches slightly higher yield than fixed refinement at lower total cost. MMR-$k$ gives the best overall portfolio tradeoff: it has the highest strong-nonduplicate yield, the best cost per strong nonduplicate, the lowest duplicate rate among successful methods, and roughly the same total cost as random-$k$.

\begin{table}[t]
\caption{Main portfolio results across 10 seeds and 10 research-ideation environments under a unified shared-candidate-pool final matrix. Cost is reported in end-to-end tokens. Raw generation and reranking alone produce no \strongnd{} ideas. Among successful methods, MMR-$k$ gives the best overall tradeoff.}
\label{tab:main_results}
\centering
\small
\begin{tabular}{lccccc}
\toprule
Method & Strong nondup. $\uparrow$ & Cost / strong $\downarrow$ & Total cost $\downarrow$ & Dup. rate $\downarrow$ & Avg. strength $\uparrow$ \\
\midrule
Raw first-$n$ & 0.00 & -- & 60.7k & \textbf{0.022} & 3.882 \\
Rerank only & 0.00 & -- & 77.4k & 0.072 & 2.743 \\
Fixed refine all & 21.80 & 5387 & 115.4k & 0.238 & \textbf{6.386} \\
Random-$k$ refinement & 22.18 & 4859 & \textbf{105.0k} & 0.186 & 5.996 \\
MMR-$k$ refinement & \textbf{23.70} & \textbf{4492} & 105.2k & \textbf{0.172} & 6.249 \\
Micro-low $\rightarrow$ heavy-$k$ & 22.40 & 5181 & 113.3k & 0.190 & 6.100 \\
\bottomrule
\end{tabular}
\end{table}

The negative controls are informative. Raw first-$n$ and rerank-only both fail on the primary metric. This suggests that the main bottleneck is not merely choosing from raw ideas. Ideas need transformation. Uniform refinement is a much stronger baseline and has the highest average research-strength score, but it is also the most expensive successful method and has the highest duplicate rate. This supports the paper's central claim: expensive refinement helps, but applying it uniformly is not the best allocation.

\subsection{Cost-yield frontier and cost decomposition}

\begin{figure}[t]
  \centering
  \includegraphics[width=0.92\linewidth]{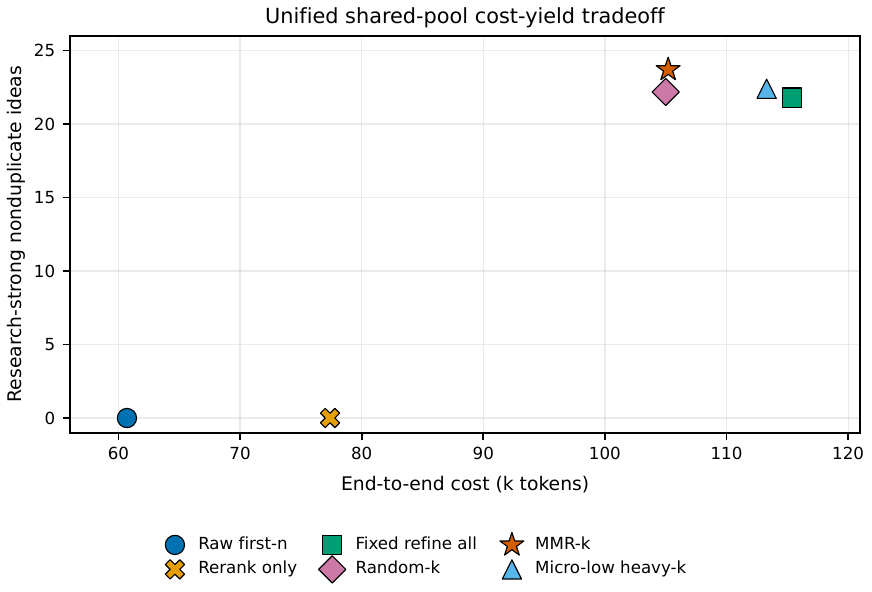}
  \caption{Cost-yield frontier for budgeted subset refinement strategies. The $x$-axis shows mean end-to-end token cost and the $y$-axis shows mean \strongnd{} ideas across 10 seeds and 10 research-ideation environments. Raw generation and reranking produce no strong nonduplicate ideas. MMR-$k$ gives the best successful tradeoff.}
  \label{fig:cost_yield_frontier}
\end{figure}

Figure~\ref{fig:cost_yield_frontier} shows the resulting cost-yield frontier. Random-$k$ lies on the low-cost successful part of the frontier: it produces slightly more strong nonduplicate ideas than fixed refinement while reducing total end-to-end cost from 115.4k to 105.0k tokens. MMR-$k$ nearly matches random-$k$ in total cost, while increasing strong-nonduplicate yield from 22.18 to 23.70 and lowering duplicate rate from 0.186 to 0.172.

\begin{figure}[t]
  \centering
  \includegraphics[width=0.78\linewidth]{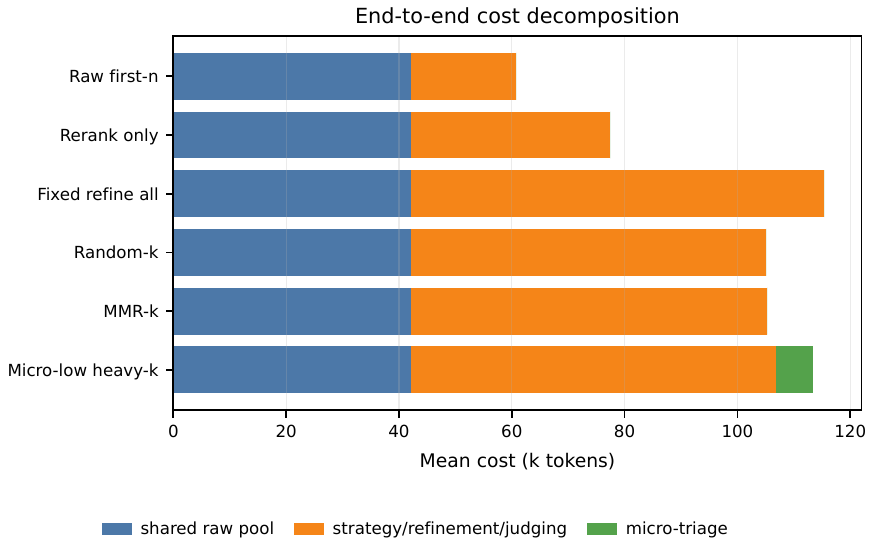}
  \caption{Mean end-to-end cost decomposition. All final strategies share the same raw candidate-pool cost. MMR-$k$ and random-$k$ reduce total cost by limiting expensive refinement, while micro-low pays an additional micro-triage cost.}
  \label{fig:cost_decomposition}
\end{figure}

The cost decomposition in Figure~\ref{fig:cost_decomposition} clarifies where savings come from. All methods share the same raw candidate-pool cost. Random-$k$ and MMR-$k$ reduce downstream cost by refining 40 selected candidates per seed rather than uniformly refining all 50 final candidates. Micro-low also refines 40 candidates, but pays an extra micro-triage cost and therefore does not match the total-cost efficiency of random-$k$ or MMR-$k$.

\subsection{Selector roles}

Table~\ref{tab:selector_roles} summarizes the role of each successful budgeted refinement policy. Random-$k$ is the low-cost successful baseline. MMR-$k$ gives the best overall tradeoff and should be treated as the main final method. Micro-low is best interpreted as a recoverable-idea triage variant: it can find strong ideas in some environments, but it is less stable and less efficient than MMR in the unified comparison.

\begin{table}[t]
\caption{Different budgeted subset policies emphasize different portfolio goals under the unified shared-pool final result.}
\label{tab:selector_roles}
\centering
\small
\begin{tabular}{lll}
\toprule
Policy & Primary role & Evidence \\
\midrule
Random-$k$ & Low-cost successful baseline & Lowest total cost among successful methods: 105.0k \\
MMR-$k$ & Best overall portfolio tradeoff & 23.7 strong nonduplicates, 4492 cost/strong, 0.172 duplicate rate \\
Micro-low $\rightarrow$ heavy-$k$ & Recoverable-idea triage variant & Competitive with fixed, but lower yield and higher cost than MMR \\
\bottomrule
\end{tabular}
\end{table}

The strength of random-$k$ is important. It shows that future work on learned or heuristic idea selection should compare against random budgeted refinement, not only against uniform refinement or raw generation. Much of the gain comes from the decision to spend expensive refinement on a subset. MMR improves on this baseline by making the subset diversity-aware rather than purely random.

\subsection{Paired comparisons}

The seed-paired tests reinforce the main result. Compared with fixed refine all, MMR produces 1.9 more strong nonduplicate ideas on average, lowers cost per strong nonduplicate by 895 tokens, and lowers duplicate rate by 0.066. MMR beats fixed refinement on cost per strong nonduplicate in 9/10 seeds and on duplicate rate in 10/10 seeds.

Compared with random-$k$, MMR produces 1.52 more strong nonduplicate ideas on average and lowers cost per strong nonduplicate by 367 tokens, while increasing total end-to-end cost by only 211 tokens on average. This is the central tradeoff: random-$k$ is a strong low-cost baseline, but MMR preserves almost the same total cost while improving yield and diversity.

Micro-low remains a secondary variant. Compared with fixed refinement, it modestly improves total cost and duplicate rate, but compared with MMR it has lower yield, higher cost per strong, higher total cost, and a higher duplicate rate. We therefore interpret micro-low as a recoverable-idea triage policy rather than the primary result.

\subsection{External-judge robustness check}
\label{sec:external_judge_validation}

Because the main benchmark uses the same open model family for generation and automatic judging, we run a blinded cross-model robustness check on a balanced 72-item validation packet sampled from the unified final run. The packet contains 12 ideas per strategy, hides strategy labels, and asks independent external judge models to score research strength, execution-readiness, novelty, feasibility, specificity, and development priority. We use this panel to test whether the broad refinement effect is visible outside the original generator/judge model family. It is not intended as a benchmark of judge models themselves.

Table~\ref{tab:external_judge_primary} summarizes the independent-family results. The average composite score of refined strategies is 6.87, compared with 5.31 for raw or reranked strategies, a 1.55-point gap on the 1-9 quality scale. This supports the broad conclusion that refinement improves proxy-rated idea quality beyond the original judge. However, the external panel does not show uniform per-item dominance by MMR. Fixed refinement has the best mean per-item rank among external judges, while MMR ranks first under two judges and lower under two others. We therefore interpret MMR's main advantage as portfolio-level: in the shared-pool benchmark, it achieves better strong-nonduplicate yield, duplicate rate, and cost efficiency, even though fixed refinement can receive stronger per-item ratings when every candidate is refined. The validation panel should be read as a robustness check for the proxy benchmark, not as a replacement for expert review or execution-grounded evaluation.

\begin{table}[t]
\caption{Blinded external-judge validation on a balanced 72-item sample. Scores are averaged over four independent-family judge models. Composite is the mean of 1-9 proxy-quality ratings. Lower mean rank is better. The panel supports the broad value of refinement, while per-item rankings among refined strategies vary by judge.}
\label{tab:external_judge_primary}
\centering
\small
\begin{tabular}{lcc}
\toprule
Strategy & Composite $\uparrow$ & Mean rank $\downarrow$ \\
\midrule
Raw first-$n$ & 5.34 & 5.50 \\
Rerank only & 5.29 & 5.50 \\
Fixed refine all & \textbf{7.01} & \textbf{1.75} \\
Random-$k$ refinement & 6.91 & 2.75 \\
MMR-$k$ refinement & 6.77 & 2.50 \\
Micro-low $\rightarrow$ heavy-$k$ & 6.77 & 3.00 \\
\bottomrule
\end{tabular}
\end{table}

\subsection{Qualitative patterns}
\label{sec:qualitative_patterns}

Qualitative inspection helps explain why selection alone fails and why diversity-aware refinement helps. Raw candidates can be thematically plausible while missing the executable details needed for a research plan. For example, a raw idea on crowdsourced feedback for quality--cost tradeoffs scored poorly because it violated the no-new-human-study constraint and lacked a concrete method. Reranking can select better-looking candidates, but it still cannot transform underspecified ideas or prevent duplicates by itself. In one rerank-only example, the highest-scoring candidate for novelty checking was still marked duplicate.

Refinement changes the character of the outputs. Fixed refinement can produce highly specific ideas, such as near-duplicate detection using title embeddings with explicit baselines and metrics, but it spends compute on every candidate and yields the highest duplicate rate among successful methods. MMR-k refinement tends to select a more varied subset before refinement. Representative MMR outputs cover evaluation-card quality, context-aware citation generation, adaptive support budgeting, and cost-effective experiment selection, rather than over-concentrating on one local cluster of novelty-checking ideas. This pattern supports the quantitative result: when refinement budget is scarce, selecting diverse candidates is not merely aesthetic. It improves the proxy-rated final portfolio.

\section{Analysis and Design Implications}

\subsection{Refinement is necessary, but not all refinement is equally useful}

The sharpest negative result is that raw generation and reranking alone produce no strong nonduplicate ideas. This mirrors concerns in recent research-ideation studies: plausible-looking proposals are not necessarily executable, and LLMs are not reliable evaluators of their own ideas \citep{si2024canllms,si2025ideationexecutiongap}. In our setting, selection without transformation is insufficient. The system must improve candidate ideas, not merely sort them.

At the same time, uniform refinement is not the best answer. Fixed refine all is strong, but it is less efficient and more duplicative than budgeted alternatives. This suggests that ideation support should be adaptive at the level of candidate ideas: some ideas need deep refinement, some need diversity-preserving selection, and some should simply be discarded.

\subsection{Why does diversity-aware subset refinement win?}

The unified result makes the role of diversity clearer than the earlier non-unified aggregate. Raw generation creates a broad but redundant candidate pool. Uniform refinement then spends expensive support on candidates even when they are semantically close to one another. Random-$k$ avoids some of this waste by refining fewer candidates, which explains why it is a strong baseline. MMR-$k$ improves the random baseline by explicitly selecting a subset that balances promise and nonredundancy.

This mechanism explains why MMR can improve both duplicate rate and strong nonduplicate yield. Diversity-aware selection does not merely make the portfolio look more varied. It changes where expensive refinement is spent. Refining four diverse candidates is often more useful than refining four near-neighbors, because only nonduplicate successes count toward the portfolio objective. This is why MMR's lowest duplicate rate and highest strong-nonduplicate yield are aligned rather than competing in the final result.

\subsection{Budgeted refinement as human-centered scaffolding}

Although this paper is an automatic evaluation study, its motivation is human-centered. A useful research ideation system should not merely generate polished text for a user. It should help a researcher explore, compare, and develop a portfolio of possibilities while preserving human agency. Budgeted subset refinement is one step toward that goal. The system provides deeper support only to selected ideas, making the final portfolio more useful for a human researcher to inspect, critique, revise, and eventually execute.

This framing connects to broader work on human-centered LLM systems and research agents. End-to-end automated research systems aim to generate ideas, implement them, execute experiments, and learn from feedback \citep{lu2024aiscientist,si2026executiongrounded}. Our work studies a lower-cost layer before full execution: improving the candidate portfolio so that human or automated executors have better starting points. From a design perspective, the result suggests that future research assistants should expose multiple refined directions, preserve portfolio diversity, and allocate more support to ideas where refinement is likely to produce a distinct executable plan.

\subsection{Scope of the proxy benchmark}

The blinded external-judge panel reduces, but does not eliminate, the evaluation-validity concern. It shows that the broad refinement effect transfers to several independent model families, while keeping the paper's claims scoped to proxy-rated proposal quality. The correct interpretation is therefore layered: the shared-pool benchmark supports relative conclusions about portfolio-level refinement-budget allocation, and the external-judge panel supports the broad per-item value of refinement under independent proxy rubrics. The paper does not claim that these proposals would necessarily produce stronger completed research artifacts after implementation.

\section{Related Work}

\paragraph{LLM research ideation.}
Recent work has evaluated whether LLMs can generate novel research ideas. Si et al. conduct a large-scale human study with over 100 NLP researchers and find that LLM-generated ideas can be judged more novel than expert ideas, while also identifying limitations in diversity and LLM self-evaluation \citep{si2024canllms}. Other research-agent systems use retrieval, iterative feedback, multi-agent collaboration, and automated review to generate scientific hypotheses or project proposals \citep{baek2025researchagent,wang2024scimon,li2025chainofideas,yang2024hypotheses}. Our work differs by focusing on refinement-budget allocation after candidate generation.

\paragraph{The ideation-execution gap.}
The most direct motivation for our work is the gap between ideas that look good before execution and ideas that survive implementation. Si et al. recruit expert researchers to execute human- and LLM-generated ideas and find that LLM ideas drop substantially more after execution across novelty, excitement, effectiveness, and overall score \citep{si2025ideationexecutiongap}. Our work studies a pre-execution intervention: can allocating refinement across a candidate portfolio improve the starting point before expensive execution?

\paragraph{LLM-as-judge and novelty evaluation.}
Because our benchmark relies on proxy scoring, it is also related to work on LLM-as-a-judge evaluation and literature-grounded novelty assessment. General LLM-as-judge studies emphasize that automatic judges can be useful but require careful calibration, bias analysis, and transparent reporting \citep{zheng2023judging,gu2024llmjudge}. Recent novelty-checking work similarly argues that scientific-idea evaluation should be grounded in retrieved prior literature rather than judged from a prompt alone \citep{shahid2025novelty}. Our external-model panel is therefore a robustness check, not a substitute for expert or execution-grounded validation.

\paragraph{Execution-grounded automated research.}
Recent systems move beyond proposal-stage evaluation by using automated executors and execution feedback. The AI Scientist and related work aim to automate parts of the research process \citep{lu2024aiscientist,yamada2025aiscientistv2}. Si et al. build automated execution environments for LLM pre-training and post-training and show that execution-guided search can outperform baselines, while reinforcement learning from execution reward can suffer diversity collapse \citep{si2026executiongrounded}. Our work is complementary: it studies budgeted refinement as a lower-cost scaffold before full execution feedback.

\paragraph{Self-refinement and critique.}
LLM systems can improve outputs through iterative feedback, critique, and revision \citep{madaan2023selfrefine,shinn2023reflexion}. In research ideation, refinement is especially important because raw ideas may be underspecified, infeasible, or redundant. Our contribution is not a new refinement prompt alone, but a comparison of policies for deciding which ideas should receive expensive refinement.

\paragraph{Diversity-aware selection.}
Maximal marginal relevance and related diversity-aware or submodular selection methods balance relevance with novelty or nonredundancy \citep{carbonell1998mmr,lin2011submodular}. We use MMR as a simple diversity-aware selector and find that it improves the overall portfolio tradeoff relative to both uniform refinement and random subset refinement in the unified final result.

\section{Limitations}

The most important limitation is evaluation validity. Our main benchmark relies on an LLM judge and automatic deduplication, while prior work shows that LLMs can be unreliable research-idea evaluators \citep{si2024canllms}. The same open model family is also used for generation and automatic judging. The blinded external-judge panel in Section~\ref{sec:external_judge_validation} partially addresses this concern by showing that the refinement effect transfers across independent model families, but it is still a proxy-evaluation check. We therefore interpret the results as controlled comparisons among allocation strategies under proxy evaluation protocols, not as final evidence that the generated ideas would succeed after implementation or expert review.

Second, ``research-strong'' does not mean experimentally validated. The execution-gap literature shows that ideas can lose apparent quality after implementation \citep{si2025ideationexecutiongap}. The benchmark should therefore be understood as a way to compare pre-execution allocation policies, not as a substitute for downstream execution studies or expert domain review.

Third, our cost metric is a token-based proxy. It captures end-to-end generation and refinement cost, but not wall-clock latency, throughput, or API pricing under different deployment settings.

Fourth, the study uses one main open model setup and a bounded set of research environments. Broader model families, domains, human-in-the-loop workflows, and execution-grounded studies are needed before drawing general conclusions about LLM research ideation.

Finally, our subset policies are intentionally simple. This is a strength for interpretability, but future work should test learned selectors, uncertainty-aware selectors, retrieval-grounded novelty estimators, and human-in-the-loop routing policies against random-$k$ and MMR-$k$ baselines.

\section{Conclusion}

We introduced \method{} for execution-aware LLM research ideation. The main finding is that useful ideation support requires more than generation or reranking under a controlled proxy benchmark: raw and reranked candidates produce no \strongnd{} ideas under our rubric, while refinement is necessary for strong proxy-rated portfolios. However, refinement should not be applied uniformly. Random subset refinement is a strong low-cost baseline, but diversity-aware MMR subset refinement gives the best overall portfolio tradeoff under a unified shared-candidate-pool evaluation: higher strong-nonduplicate yield, lower duplicate rate, lower cost per strong nonduplicate, and lower total cost than fixed refinement. A blinded external-judge panel supports the broad value of refinement while reinforcing that MMR's advantage is portfolio-level rather than uniform per-item dominance. These results suggest that LLM research ideation systems should be understood as scaffolding systems that allocate support across candidate ideas, while leaving downstream scientific value to expert judgment and execution-grounded validation.

\section*{Broader Impact Statement}

This work aims to improve LLM-assisted research ideation as a form of cognitive scaffolding for difficult knowledge work. More effective ideation support could help researchers explore more possibilities, refine weak ideas, and identify promising directions earlier. However, more capable ideation systems could also increase the volume of low-quality or automated submissions if used without human judgment and execution-grounded validation. We therefore frame budgeted subset refinement as a support mechanism for human researchers, not as a substitute for expert evaluation, responsible experimentation, or scientific accountability.

\section*{AI Assistance Disclosure}
The author used AI-assisted tools for implementation support, debugging, analysis checks, and manuscript editing. The author reviewed, modified, and takes responsibility for all code, experiments, analyses, interpretations, and claims.

\bibliographystyle{plainnat}
\bibliography{paper2_references}

\appendix

\section{Full Experimental Configuration}
\label{app:config}

The final unified matrix uses \texttt{Qwen/Qwen2.5-3B-Instruct} as both generator and judge, bfloat16 CUDA inference, 10 random seeds, and 10 research-ideation environments. Each environment begins with 16 raw candidates. Final portfolios contain 50 ideas per strategy group aggregated across the 10 environments for one seed. The evaluated strategy groups are raw first-$n$, rerank-only, fixed refine all, random-$k$ refinement, MMR-$k$ refinement, and micro-low $\rightarrow$ heavy-$k$ refinement. The random-$k$ result averages five random subset draws per seed. Internal run identifiers are used in the project code and result CSVs, but the paper uses descriptive strategy names throughout.

Generation temperature is 0.9 for cheap raw generation, 0.7 for expensive generation, and 0.6 for heavy refinement. Heavy refinement uses up to 768 new tokens. Single judge calls use up to 256 new tokens and batched judging uses up to 1536 new tokens. The high-score threshold is 6.0. Deduplication uses \texttt{all-MiniLM-L6-v2} embeddings with a near-duplicate threshold of 0.82.

\section{Algorithmic Summary}
\label{app:algorithm}

The following pseudocode summarizes the shared-pool evaluation used by all final strategies. It is intended to make the allocation comparison explicit. Appendix~\ref{app:prompts} gives compact prompt templates, and Appendix~\ref{app:compute_artifacts} summarizes compute resources and external assets.

\begin{enumerate}[leftmargin=1.5em]
    \item For each seed and research environment, generate one shared raw candidate pool $\mathcal{C}_e$ of 16 ideas.
    \item For each strategy $s$, choose a strategy-specific selected set $S_e^s$: raw first-$n$ keeps raw candidates, rerank-only selects without refinement, fixed refinement sets $S_e^s=\mathcal{C}_e$, random-$k$ samples $k=4$, MMR-$k$ applies Equation~\ref{eq:mmr}, and micro-low applies Equation~\ref{eq:microlow}.
    \item Apply heavy refinement only to the candidates selected for refinement by $s$.
    \item Score all final strategy outputs with the execution-aware judge rubric.
    \item Deduplicate each final portfolio using sentence-embedding similarity and count only research-strong nonduplicate ideas.
    \item Aggregate yield, cost, duplicate rate, and paired differences across the same 10 seed indices.
\end{enumerate}

\section{Prompts and Rubrics}
\label{app:prompts}

All strategies share the same research-environment framing, which specifies the problem, allowed data or corpora, allowed baselines, allowed metrics, constraints, forbidden approaches, and expected artifact. This appendix gives compact templates so that the experimental interface is auditable. The implementation uses these templates with environment-specific fields filled from the benchmark configuration.

\paragraph{Research-environment framing.}
\begin{prompttemplate}
\textbf{Fields.} Environment name, goal, allowed data or corpora, allowed baselines, allowed metrics, constraints, forbidden approaches, and expected artifact.\\
\textbf{Instruction.} Propose concrete, execution-aware ML/NLP research ideas that can be developed as a 1-2 week project plan with a method, data source, baseline, metric, expected result, and likely failure mode.
\end{prompttemplate}

\paragraph{Raw generation prompt.}
\begin{prompttemplate}
Generate diverse research ideas for the specified environment. Each idea must include a title, research question, method, dataset or corpus, baseline, metric, expected result, implementation plan, and likely failure mode. Prefer specific, feasible, low-resource ideas over vague product concepts. Avoid new human-subject studies unless explicitly allowed. Return structured records.
\end{prompttemplate}

\paragraph{Heavy refinement prompt.}
\begin{prompttemplate}
Given a candidate idea, refine it into a more specific and execution-ready research proposal. Strengthen the research question, mechanism, dataset or corpus, baseline, metric, expected result, implementation plan, feasibility assumptions, and failure modes. Do not merely polish wording: if the idea is vague, make it concrete. If it is too broad, narrow it to a runnable 1-2 week study. Return the same structured fields as the raw idea.
\end{prompttemplate}

\paragraph{Micro-triage prompt.}
\begin{prompttemplate}
Given a rough candidate idea, cheaply assess whether it is a low-scoring but recoverable idea that deserves heavy refinement. Score recoverability, missing execution details, feasibility, and likely improvement after refinement. Return a compact structured output with a recoverability score and brief rationale.
\end{prompttemplate}

\paragraph{Automatic judge rubric.}
\begin{prompttemplate}
Score each idea on novelty, excitement, feasibility, expected effectiveness, specificity, execution-readiness, dataset specificity, baseline specificity, metric specificity, resource feasibility, implementation clarity, human-subject dependency, and method sanity. The aggregate research-strength score combines these dimensions and penalizes vague methods, weak metrics, excessive human-subject dependence, unrealistic compute requirements, and proposals that restate the benchmark goal without an executable mechanism.
\end{prompttemplate}

\paragraph{External-judge validation prompt.}
\begin{prompttemplate}
Act as an independent research-idea evaluator who does not know which method produced the idea. Rate the idea as a potential 1-2 week ML/NLP research project for one independent researcher with a local GPU or modest API budget. Use 1-9 ratings for research strength, execution-readiness, novelty, feasibility, and specificity, and a 1-5 rating for development priority. Be strict about vague methods, missing baselines or metrics, expensive human-subject studies, unrealistic compute, and ideas that merely restate a system goal. Return only the requested structured rating fields and a brief rationale.
\end{prompttemplate}

\section{Compute Resources, External Assets, and Licenses}
\label{app:compute_artifacts}

\paragraph{Hardware.}
The final experiments and validation runs were conducted locally on a single NVIDIA RTX 3090 GPU with 24GB of VRAM. Larger external judge models were loaded with 4-bit quantization when needed to fit on the device. Exact wall-clock runtime and GPU-hours were not logged systematically. The paper therefore reports token-based end-to-end cost as the primary compute proxy.

\paragraph{Models and software.}
The main generator and automatic judge is \texttt{Qwen/Qwen2.5-3B-Instruct} \citep{qwen2025technical}. We record external-asset licenses here for transparency. The public repository metadata lists Qwen2.5-3B-Instruct under \texttt{qwen-research}, Phi-4-mini-instruct under MIT, Llama-3.1-8B-Instruct under the Llama 3.1 Community License, Mistral-7B-Instruct-v0.3 under Apache-2.0, Granite-3.3-8B-Instruct under Apache-2.0, and all-MiniLM-L6-v2 under Apache-2.0. The external validation panel uses Phi-4-mini-instruct, Llama-3.1-8B-Instruct, Mistral-7B-Instruct-v0.3, and Granite-3.3-8B-Instruct. Deduplication uses the all-MiniLM-L6-v2 sentence embedding model \citep{reimers2019sentencebert}. The implementation uses standard Python ML tooling, including PyTorch, Transformers, sentence-transformers, pandas, NumPy, and scikit-learn-compatible analysis utilities.

\paragraph{Data and human subjects.}
The benchmark environments and candidate ideas are generated research-proposal artifacts rather than human-subject data. The paper does not collect new human ratings or conduct a human-subject study. The external-judge validation packet is a blinded sample of generated ideas used for the cross-model robustness check.

\paragraph{Reporting details.}
The main text and appendices report the paper's claims and scope (Sections~1 and~8), experimental configuration (Appendix~\ref{app:config}), algorithmic structure (Appendix~\ref{app:algorithm}), prompts and rubrics (Appendix~\ref{app:prompts}), statistics and paired comparisons (Sections~4-5 and Appendix~\ref{app:paired}), compute resources, external assets, licenses, and human-subject status. No new human-subject data are collected.

\section{Threshold Sensitivity}
\label{app:threshold}

Table~\ref{tab:threshold_sensitivity} recomputes the primary yield metric at several research-strength thresholds while preserving the same deduplication assignments. MMR remains the highest-yield successful refinement method at thresholds 5.5, 6.0, 6.5, and 7.0. Raw generation and reranking remain ineffective at the paper's main threshold of 6.0 and above.

\begin{table}[H]
\caption{Threshold sensitivity: mean nonduplicate ideas above each research-strength threshold.}
\label{tab:threshold_sensitivity}
\centering
\small
\begin{tabular}{lcccccc}
\toprule
Threshold & Raw & Rerank & Fixed & Random-$k$ & MMR-$k$ & Micro-low \\
\midrule
5.5 & 3.90 & 1.40 & 23.70 & 24.26 & \textbf{25.80} & 23.80 \\
6.0 & 0.00 & 0.00 & 21.80 & 22.18 & \textbf{23.70} & 22.40 \\
6.5 & 0.00 & 0.00 & 20.50 & 20.30 & \textbf{21.60} & 20.90 \\
7.0 & 0.00 & 0.00 & 17.50 & 17.24 & \textbf{18.70} & 17.00 \\
\bottomrule
\end{tabular}
\end{table}

\begin{figure}[H]
  \centering
  \includegraphics[width=0.88\linewidth]{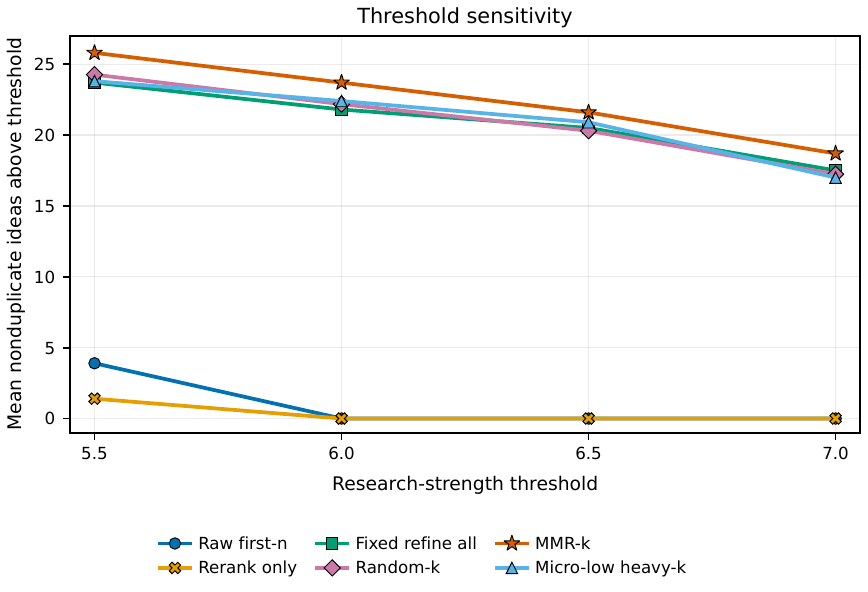}
  \caption{Threshold sensitivity for strong nonduplicate yield. MMR-$k$ remains the highest-yield successful method across the tested thresholds.}
\end{figure}

\section{Per-Seed and Per-Topic Results}
\label{app:perseed}

Table~\ref{tab:per_seed_strong} reports per-seed \strongnd{} counts for the four successful refinement methods. Raw first-$n$ and rerank-only produce zero \strongnd{} ideas in every seed.

\begin{table}[H]
\caption{Per-seed \strongnd{} counts for successful refinement methods.}
\label{tab:per_seed_strong}
\centering
\small
\begin{tabular}{ccccc}
\toprule
Seed & Fixed & Random-$k$ & MMR-$k$ & Micro-low \\
\midrule
0 & 23 & 23.6 & 23 & 21 \\
1 & 26 & 23.4 & 24 & 24 \\
2 & 22 & 21.4 & 23 & 18 \\
3 & 22 & 24.8 & 23 & 17 \\
4 & 20 & 23.4 & 26 & 25 \\
5 & 23 & 20.6 & 24 & 28 \\
6 & 18 & 22.2 & 26 & 22 \\
7 & 22 & 21.2 & 18 & 24 \\
8 & 21 & 22.0 & 27 & 19 \\
9 & 21 & 19.2 & 23 & 26 \\
\bottomrule
\end{tabular}
\end{table}

Table~\ref{tab:per_topic} reports per-topic mean \strongnd{} counts for successful refinement methods. No single policy dominates every environment, which supports interpreting the result as a portfolio-allocation finding rather than as proof of a universally optimal selector.

\begin{table}[H]
\caption{Per-topic mean \strongnd{} counts over 10 seeds. Topic labels are shortened for readability. The full environment descriptions are summarized in Appendix~\ref{app:config}.}
\label{tab:per_topic}
\centering
\scriptsize
\setlength{\tabcolsep}{3.8pt}
\begin{tabular}{lcccc}
\toprule
Topic & Fixed & Random-$k$ & MMR-$k$ & Micro-low \\
\midrule
Adaptive support budgeting & 1.50 & 1.96 & 2.00 & \textbf{2.40} \\
Agent evaluation cards & \textbf{2.80} & 1.58 & 2.20 & 2.10 \\
Cognitive scaffolding tasks & \textbf{2.10} & 1.66 & 1.80 & 1.60 \\
Collaborative agent evaluation & 3.10 & 2.78 & \textbf{3.30} & 2.90 \\
Duplicate-aware ideation & 2.80 & 2.88 & 2.60 & \textbf{3.00} \\
Execution-readiness prediction & 1.10 & 2.58 & \textbf{2.90} & 1.40 \\
Execution-trace selection & 1.40 & 1.16 & 1.30 & \textbf{2.30} \\
Ideation quality-cost & 2.40 & 2.66 & 2.60 & \textbf{3.30} \\
Novelty-checking pipeline & \textbf{2.40} & 1.72 & 1.90 & 2.10 \\
Research writing assistance & 2.20 & \textbf{3.20} & 3.10 & 1.30 \\
\bottomrule
\end{tabular}
\end{table}

\section{Additional Paired Comparisons}
\label{app:paired}

Table~\ref{tab:paired} summarizes selected seed-paired comparisons. Positive yield differences favor the first method; negative cost and duplicate-rate differences favor the first method.

\begin{table}[H]
\caption{Selected paired comparisons over 10 seeds.}
\label{tab:paired}
\centering
\small
\begin{tabular}{llccc}
\toprule
Comparison & Metric & Mean diff. & 95\% CI & Win count \\
\midrule
MMR $-$ fixed & Strong nondup. & +1.90 & [0.20, 3.60] & 7/10 \\
MMR $-$ fixed & Cost / strong & -895 & [-1327, -498] & 9/10 \\
MMR $-$ fixed & Duplicate rate & -0.066 & [-0.090, -0.042] & 10/10 \\
MMR $-$ random & Strong nondup. & +1.52 & [-0.22, 2.82] & 9/10 \\
MMR $-$ random & Cost / strong & -367 & [-667, 71] & 9/10 \\
MMR $-$ micro-low & Cost / strong & -690 & [-1325, -45] & 7/10 \\
Random $-$ fixed & Cost / strong & -528 & [-918, -134] & 7/10 \\
Micro-low $-$ fixed & Cost / strong & -205 & [-839, 504] & 6/10 \\
\bottomrule
\end{tabular}
\end{table}

\section{Cost Decomposition}
\label{app:cost}

Table~\ref{tab:cost_decomp} reports the same cost decomposition shown in Figure~\ref{fig:cost_decomposition}. All methods share the same raw candidate-pool cost. The main cost differences arise from downstream refinement, judging, and micro-triage.

\begin{table}[H]
\caption{Mean cost decomposition in thousands of tokens. Downstream cost includes strategy-specific refinement, routing, and judging after the shared raw pool.}
\label{tab:cost_decomp}
\centering
\small
\begin{tabular}{lccccc}
\toprule
Method & Raw pool & Micro pool & Downstream & End-to-end & Refined count \\
\midrule
Raw first-$n$ & 42.2 & 0.0 & 18.5 & 60.7 & 0 \\
Rerank only & 42.2 & 0.0 & 35.3 & 77.4 & 0 \\
Fixed refine all & 42.2 & 0.0 & 73.2 & 115.4 & 50 \\
Random-$k$ refinement & 42.2 & 0.0 & 62.9 & 105.0 & 40 \\
MMR-$k$ refinement & 42.2 & 0.0 & 63.1 & 105.2 & 40 \\
Micro-low $\rightarrow$ heavy-$k$ & 42.2 & 6.6 & 64.6 & 113.3 & 40 \\
\bottomrule
\end{tabular}
\end{table}

\section{Qualitative Examples}
\label{app:examples}

Table~\ref{tab:qual_examples} gives representative examples illustrating the patterns discussed in Section~\ref{sec:qualitative_patterns}. The examples are included to make the automatic metrics more interpretable, not as independent validation evidence. We choose examples with concrete titles and explicit research questions so the reader can inspect the mechanism without relying only on aggregate scores.

\begin{table}[H]
\caption{Representative qualitative examples from the unified final run.}
\label{tab:qual_examples}
\centering
\scriptsize
\begin{sloppypar}
\renewcommand{\arraystretch}{1.08}
\resizebox{\linewidth}{!}{%
\begin{tabular}{>{\raggedright\arraybackslash}p{0.13\linewidth}>{\raggedright\arraybackslash}p{0.24\linewidth}>{\raggedright\arraybackslash}p{0.34\linewidth}>{\raggedright\arraybackslash}p{0.23\linewidth}}
\toprule
Pattern & Example title & Research question & Interpretation \\
\midrule
Raw failure & Quality-Cost Evaluation via Corpus Analysis & How can we evaluate LLM-generated ideas in terms of quality and cost using a literature corpus? & Low-scoring raw idea. It repeats the benchmark framing but lacks a concrete method, baseline, and executable measurement plan. \\
Rerank failure & Early Signal Analysis for Selecting Executable Plans & Can we use cheap pre-execution signals to predict which LLM-generated experiment plans will lead to successful executions? & Reranking can select a plausible candidate, but the output remains near-duplicate and underspecified without refinement. \\
Fixed success & Detection of Near-Duplicate Research Ideas Using Title Embeddings & Can we accurately detect AI-generated research ideas that are too close to existing literature? & Uniform refinement can produce very strong ideas, but it spends the same effort on many candidates and has the highest duplicate rate. \\
MMR success & Adaptive Budgeting for High-Stakes Support Responses & How does dynamic allocation of LLM resources affect the quality and cost of support responses in high-stakes scenarios? & MMR selects a diverse high-quality candidate outside the dominant novelty-checking cluster. \\
MMR coverage & Cost-Effective Experiment Selection for LLM-generated Plans & What criteria can be used to select LLM-generated experiment plans with the lowest expected cost per executable plan? & MMR adds portfolio coverage of execution-planning topics while maintaining strong specificity. \\
Micro-low success & Optimizing Execution Plan Selection for Cheap Pre-Execution Signals & How can we effectively select LLM-generated experiment plans for execution using cheap pre-execution signals? & Micro-low sometimes recovers strong ideas from cheap triage, but is not as stable as MMR overall. \\
\bottomrule
\end{tabular}%
}
\renewcommand{\arraystretch}{1.0}
\end{sloppypar}
\end{table}

\section{Validation Packet and External-Model Robustness Protocol}
\label{app:validation_packet}

The external-model robustness check uses a balanced, blinded validation packet generated from the unified final run. The external-judge prompt template is shown in Appendix~\ref{app:prompts}. Each sampled idea hides its generating strategy and includes the research environment, title, research question, method summary, dataset or corpus, baseline, and metric. The packet contains 72 items, with 12 items per strategy group. The rating form asks external judge models to score research strength, execution-readiness, novelty, feasibility, specificity, and development priority, with duplicate judgments computed within small portfolios.

We use this packet for the cross-model robustness check in Section~\ref{sec:external_judge_validation}. The same blinded packet was evaluated with a four-model independent-family panel: Phi-4-mini, Llama-3.1-8B, Mistral-7B-Instruct-v0.3, and Granite-3.3-8B. These ratings are reported only as a robustness check for the broad refined-versus-unrefined gap. Additional evaluator runs were used only as internal diagnostics and are not part of the paper's headline validation result.

\paragraph{External-judge robustness protocol.}
The paper reports only the pre-specified primary validation summary rather than treating evaluator implementation details as scientific results. The scoring protocol, prompt template, and aggregate validation table are included here so that the check can be inspected from the paper itself.

\end{document}